\documentclass[letterpaper]{article} 
\usepackage{aaai23}  
\usepackage{times}  
\usepackage{helvet}  
\usepackage{courier}  
\usepackage[hyphens]{url}  
\usepackage{graphicx} 
\usepackage{natbib}  
\usepackage{caption} 
\usepackage{algorithm}
\usepackage{algorithmic}

\usepackage{newfloat}
\usepackage{listings}
\DeclareCaptionStyle{ruled}{labelfont=normalfont,labelsep=colon,strut=off} 
\floatstyle{ruled}
\newfloat{listing}{tb}{lst}{}
\floatname{listing}{Listing}
\usepackage{nameref}

\title{Demo Alleviate: \underline{Demo}nstrating \underline{A}rtificial Inte\underline{ll}igence \underline{E}nabled \underline{Vi}rtual \underline{A}ssistance for \underline{Te}lehealth: The Mental Health Case}
\author{
    Kaushik Roy, 
    Vedant Khandelwal, 
    Raxit Goswami, 
    Nathan Dolbir, 
    Jinendra Malekar, 
    Amit Sheth
}
\affiliations{
    Artificial Intelligence Institute, University of South Carolina\\ Columbia, South Carolina (Zip - 29208)\\

    {\{kaushikr, vedant, rgoswami, ndolbir\}}@email.sc.edu, jmalekar@mailbox.sc.edu, amit@sc.edu
}

\usepackage{bibentry}

\begin{document}

\maketitle

\begin{abstract}
After the pandemic, artificial intelligence (AI) powered support for mental health care has become increasingly important. The breadth and complexity of significant challenges required to provide adequate care involve:
(a) Personalized patient understanding, (b) Safety-constrained and medically validated chatbot patient interactions, and (c) Support for continued feedback-based refinements in design using chatbot-patient interactions. 
We propose Alleviate, a chatbot designed to assist patients suffering from mental health challenges with personalized care and assist clinicians with understanding their patients better. Alleviate draws from an array of publicly available clinically valid mental-health texts and databases, allowing Alleviate to make medically sound and informed decisions. In addition, Alleviate's modular design and explainable decision-making lends itself to robust and continued feedback-based refinements to its design. In this paper, we explain the different modules of Alleviate and submit a short video demonstrating Alleviate's capabilities to help patients and clinicians understand each other better to facilitate optimal care strategies. 
\end{abstract}

\section{Introduction}
The current pandemic has over-extended mental healthcare systems and caused striking increases in mental health clinical services\cite{who_art,selfcare2020}. With the severe shortage of mental health clinicians coupled with a decrease in in-person visits at health care facilities, AI-powered chatbots offer a promising solution in helping patients mitigate mental health symptoms early on through active self-care for effective prevention and intervention. The current standard of chatbots provides script-based screening tasks (e.g., reminding, scheduling) that assist patients with mental health self-management through chatbot-patient interactions for their daily self-care\cite{jaimini2018khealth}.

Enabling more advanced capabilities in chatbots raises challenging core algorithmic issues on:
(a) Personalized patient understanding, (b) Safety-constrained and medically validated chatbot-patient interactions, and (c) support for continued feedback-based refinements in design using chatbot-patient and chatbot-clinician interactions.

\begin{figure}[!h]
    \centering
    \includegraphics[width=\linewidth]{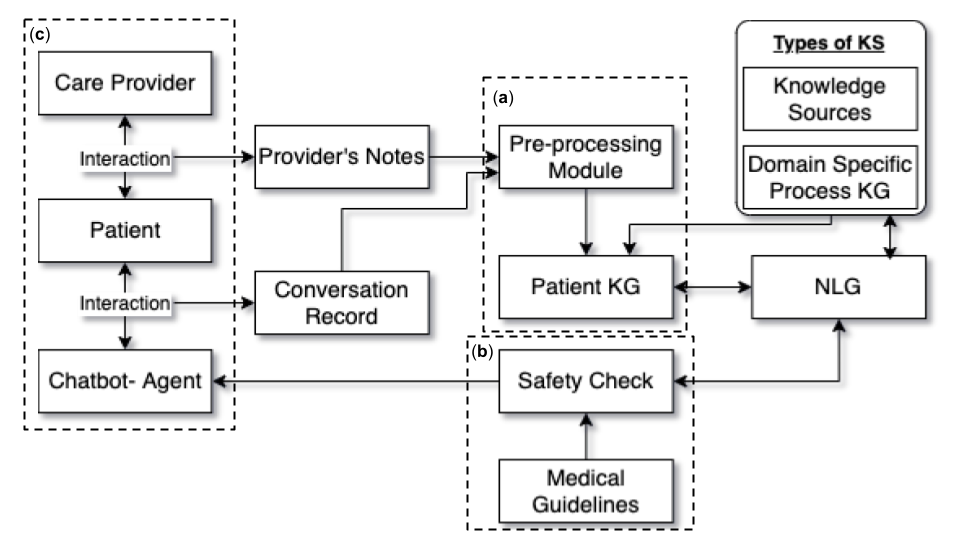}
    \caption{(a) Alleviate constructs a consolidated knowledge base by drawing from knowledge databases that are mental health domain specific - Eg: Suicide and Depression Rating scales, broader medical context based - Eg: Medication interactions and side-effects. Alleviate integrates the extracted knowledge with patient-specific information to form a personalized patient knowledge graph. (b) Alleviate's task executions conform strictly to clinically established safety standards and medical guidelines provided to Alleviate's AI backend in the form of knowledge graph path constraints. (c) Alleviate's algorithms support constant feedback-based refinements through continued patient and care-provider interactions in a reinforcement learning setup.}
    \label{fig:overview}
\end{figure}

We propose Alleviate, a chatbot designed to assist patients suffering from mental health challenges with personalized care. Alleviate represents personalized patient knowledge as a graph that integrates knowledge from an array of clinically valid mental-health texts and databases with patient-specific information derived from provider notes and patient-chatbot interactions (see Figure \ref{fig:overview} (a))\cite{cameron2015context,roy2021depression,rawte2022tdlr,lokala2021knowledge,gaur2021can}. Furthermore, alleviate operates in strict conformance with medically established guidelines ensuring safe interactions with the patient. The breadth and depth of medical knowledge consolidated in the knowledge graph enable Alleviate to make medically sound and informed decisions (see Figure \ref{fig:overview} (b))\cite{roy2022process,sheth2022process,gupta2022learning}. In addition, Alleviate's modular design and explainable reinforcement learning algorithms allow continued development and refinement using user and clinician feedback (see Figure \ref{fig:overview} (c))\cite{roy2021knowledge}. We explain the inner workings of the Alleviate functions:
\begin{itemize}
\item \nameref{sec:f1}. 
\item \nameref{sec:f2}.
\item \nameref{sec:f3}.
\end{itemize}. The functions cover Alleviate's aim to assist care providers with safe and explainable personalized patient care. 

\section{Safe and Explainable Medication Reminder and Troubleshooting}\label{sec:f1}
\begin{figure}[!h]
    \centering
    \includegraphics[width=\linewidth]{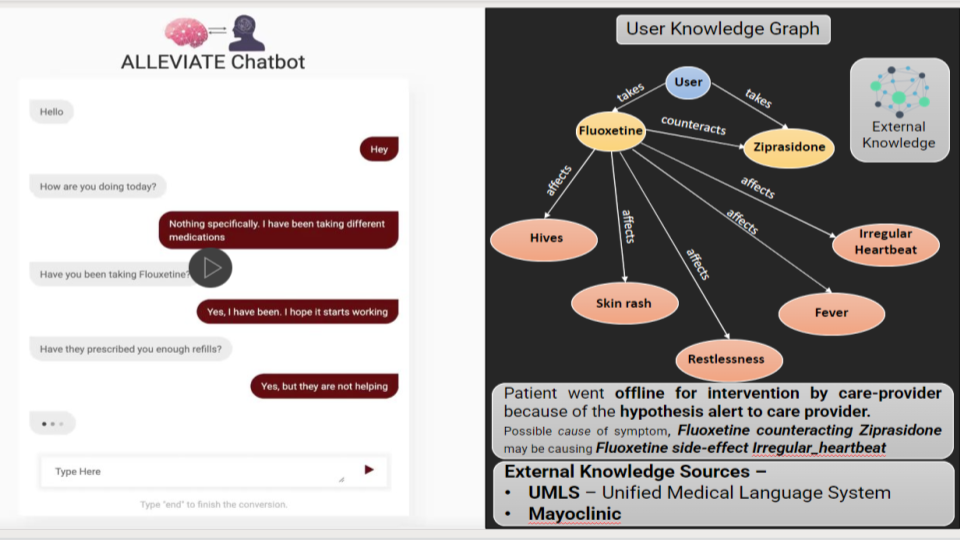}
    \caption{Alleviate integrates the user's personal medication information and the information contained in medical knowledge databases such as the mayo clinic and the Unified Medical Language System (UMLS) to perform medication inquiries and troubleshooting.}
    \label{fig:mrts}
\end{figure}

Alleviate extracts personalized patient information from provider notes and past patient interactions using <subject, predicate, object> triple extraction techniques to bootstrap the patient knowledge graph. Further, Alleviate integrates patient information with mental health information from knowledge bases by connecting the entities and relationships in the initialized patient knowledge graph with similar entities in the knowledge bases. Computing dense representation-based distances are used to determine similar entities. Finally, alleviate resolves connection conflicts during integration using clinician-specified guidelines for conflict resolution. Figure \ref{fig:mrts} Illustrates how Alleviate can also construct potential hypotheses utilizing the information from its knowledge sources (stored on a back-end server and not visible to the user). Alleviate's theories provide valuable insight to the clinician care provider.

\section{Patient Appraisal on Adherence to Medical Recommendations}\label{sec:f2}
\begin{figure}[!h]
    \centering
    \includegraphics[width=\linewidth]{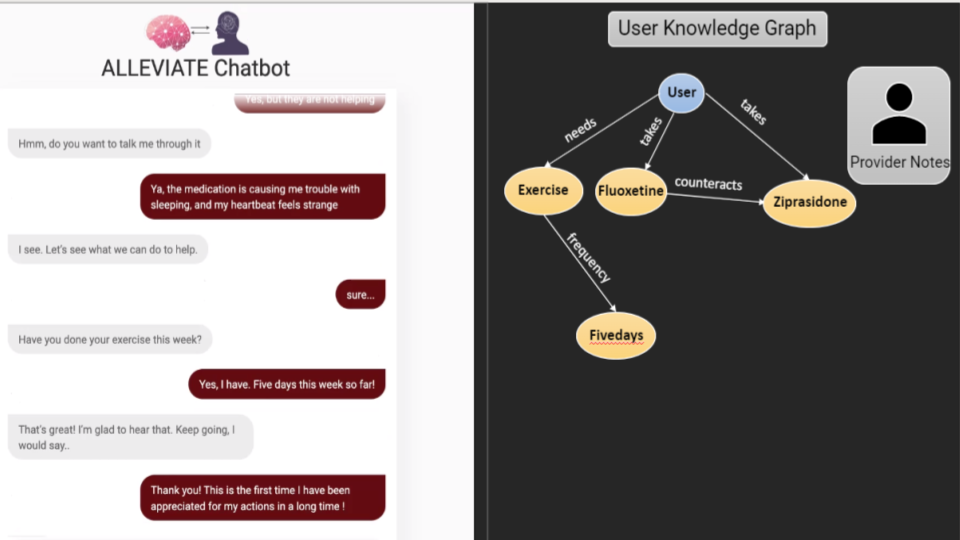}
    \caption{Alleviate praises the user for adherence to medical recommendations contained in the provider notes written by the care provider. Here, Alleviate appreciates the user accomplishing five days of exercise that week.}
    \label{fig:appraise}
\end{figure}

Alleviate's patient knowledge graph is utilized to perform inquiries about adherence to medical recommendations obtained from the provider notes written by the care provider during offline patient-provider interactions. Figure \ref{fig:appraise} shows Alleviate praising a user for completing the recommended amount of weekly exercise.
\section{Behavior Detection Requiring Emergency Human Intervention}\label{sec:f3}
Alleviate continuously performs safety checks to detect conversation patterns that require emergency human intervention. Alleviate computes dense representation similarities matching with concepts from clinically established alarming behavior detection questionnaires represented as trees to determine the time for emergency intervention. 
\begin{figure}[!h]
    \centering
    \includegraphics[width=\linewidth]{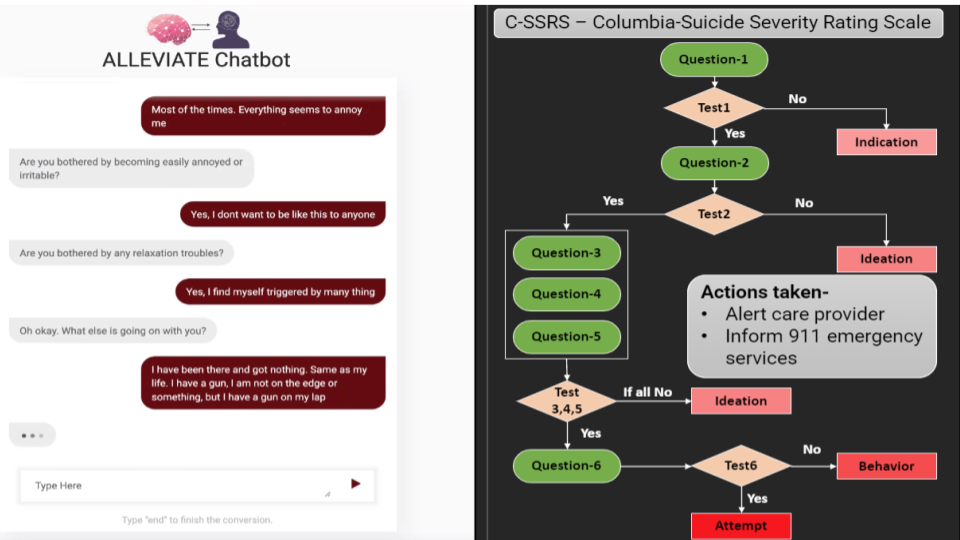}
    \caption{Alleviate constantly monitors patient conversation for patterns requiring emergency human intervention. Here, Alleviate alerts emergency services of the patient's potential suicidal ideation.}
    \label{fig:cssrs}
\end{figure}
\section{Conclusion}\label{sec:conc}
In this work, we propose Alleviate, a mental health chatbot designed to assist care providers with safe and explainable personalized patient care. Alleviate's integrated use of personal information, medical knowledge, and mental-health questionnaires encoded as graphs and trees allow easy modeling of safety conformance using graph and tree path constraints. The structure of the graphs and trees enables explanation of Alleviate's functions.

\noindent\textbf{Acknowledgements:} This research is supported by National Science Foundation (NSF) Award \# 2133842 “EAGER: Advancing Neuro-symbolic AI with Deep Knowledge-infused Learning,” \cite{sheth2019shades}. Any opinions, findings, and conclusions or recommendations expressed in this material are those of the author(s) and do not necessarily reflect the views of the NSF. We want to extend our thanks to the team of Dr. Meera Narasimhan and SHAIP\footnote{\url{https://www.shaip.com}}\footnote{\url{https://doctors.prismahealth.org/provider/Meera+Narasimhan/992922}} for providing us the data which was used for experimentation purpose in the proposed system.
\bibliography{aaai23}

\begin{thebibliography}{13}
\providecommand{\natexlab}[1]{#1}

\bibitem[{Cameron et~al.(2015)Cameron, Kavuluru, Rindflesch, Sheth,
  Thirunarayan, and Bodenreider}]{cameron2015context}
Cameron, D.; Kavuluru, R.; Rindflesch, T.~C.; Sheth, A.~P.; Thirunarayan, K.;
  and Bodenreider, O. 2015.
\newblock Context-driven automatic subgraph creation for literature-based
  discovery.
\newblock \emph{Journal of biomedical informatics}, 54: 141--157.

\bibitem[{Gaur et~al.(2021)Gaur, Roy, Sharma, Srivastava, and
  Sheth}]{gaur2021can}
Gaur, M.; Roy, K.; Sharma, A.; Srivastava, B.; and Sheth, A. 2021.
\newblock “Who can help me?”: Knowledge Infused Matching of Support Seekers
  and Support Providers during COVID-19 on Reddit.
\newblock In \emph{2021 IEEE 9th International Conference on Healthcare
  Informatics (ICHI)}, 265--269. IEEE.

\bibitem[{Gupta et~al.(2022)Gupta, Agarwal, Gaur, Roy, Narayanan, Kumaraguru,
  and Sheth}]{gupta2022learning}
Gupta, S.; Agarwal, A.; Gaur, M.; Roy, K.; Narayanan, V.; Kumaraguru, P.; and
  Sheth, A. 2022.
\newblock Learning to automate follow-up question generation using process
  knowledge for depression triage on reddit posts.
\newblock \emph{arXiv preprint arXiv:2205.13884}.

\bibitem[{Jaimini et~al.(2018)Jaimini, Thirunarayan, Kalra, Venkataraman,
  Kadariya, Sheth et~al.}]{jaimini2018khealth}
Jaimini, U.; Thirunarayan, K.; Kalra, M.; Venkataraman, R.; Kadariya, D.;
  Sheth, A.; et~al. 2018.
\newblock “How Is My Child’s Asthma?” digital phenotype and actionable
  insights for pediatric asthma.
\newblock \emph{JMIR pediatrics and parenting}, 1(2): e11988.

\bibitem[{Lokala et~al.(2021)Lokala, Gaur, Roy, and
  Sheth}]{lokala2021knowledge}
Lokala, U.; Gaur, M.; Roy, K.; and Sheth, A. 2021.
\newblock Knowledge infused Natural Language understanding for Public Health,
  Epidemiology, and Substance Use.

\bibitem[{Rawte et~al.(2022)Rawte, Chakraborty, Roy, Gaur, Faldu, Kikani,
  Akbari, and Sheth}]{rawte2022tdlr}
Rawte, V.; Chakraborty, M.; Roy, K.; Gaur, M.; Faldu, K.; Kikani, P.; Akbari,
  H.; and Sheth, A. 2022.
\newblock TDLR: Top (Semantic)-Down (Syntactic) Language Representation.
\newblock \emph{UMBC Faculty Collection}.

\bibitem[{Roy et~al.(2022)Roy, Gaur, Zhang, and Sheth}]{roy2022process}
Roy, K.; Gaur, M.; Zhang, Q.; and Sheth, A. 2022.
\newblock Process Knowledge-infused Learning for Suicidality Assessment on
  Social Media.
\newblock \emph{arXiv preprint arXiv:2204.12560}.

\bibitem[{Roy et~al.(2021{\natexlab{a}})Roy, Lokala, Khandelwal, and
  Sheth}]{roy2021depression}
Roy, K.; Lokala, U.; Khandelwal, V.; and Sheth, A. 2021{\natexlab{a}}.
\newblock " Is depression related to cannabis?": A knowledge-infused model for
  Entity and Relation Extraction with Limited Supervision.
\newblock \emph{arXiv preprint arXiv:2102.01222}.

\bibitem[{Roy et~al.(2021{\natexlab{b}})Roy, Zhang, Gaur, and
  Sheth}]{roy2021knowledge}
Roy, K.; Zhang, Q.; Gaur, M.; and Sheth, A. 2021{\natexlab{b}}.
\newblock Knowledge infused policy gradients with upper confidence bound for
  relational bandits.
\newblock In \emph{Machine Learning and Knowledge Discovery in Databases.
  Research Track: European Conference, ECML PKDD 2021, Bilbao, Spain, September
  13--17, 2021, Proceedings, Part I 21}, 35--50. Springer.

\bibitem[{Sheth et~al.(2019)Sheth, Gaur, Kursuncu, and
  Wickramarachchi}]{sheth2019shades}
Sheth, A.; Gaur, M.; Kursuncu, U.; and Wickramarachchi, R. 2019.
\newblock Shades of knowledge-infused learning for enhancing deep learning.
\newblock \emph{IEEE Internet Computing}, 23(6): 54--63.

\bibitem[{Sheth et~al.(2022)Sheth, Gaur, Roy, Venkataraman, and
  Khandelwal}]{sheth2022process}
Sheth, A.; Gaur, M.; Roy, K.; Venkataraman, R.; and Khandelwal, V. 2022.
\newblock Process Knowledge-Infused AI: Toward User-Level Explainability,
  Interpretability, and Safety.
\newblock \emph{IEEE Internet Computing}, 26(5): 76--84.

\bibitem[{WCVB(2020)}]{selfcare2020}
WCVB. 2020.
\newblock 40\% of US adults reported struggling with mental health or substance
  abuse, according to June CDC survey.
\newblock \url{http://bit.ly/news_mental_health}.
\newblock Accessed: 2022-12-10.

\bibitem[{WHO(2022)}]{who_art}
WHO. 2022.
\newblock Covid-19 pandemic triggers 25\% increase in prevalence of anxiety and
  depression worldwide.
\newblock
  \url{https://www.who.int/news/item/02-03-2022-covid-19-pandemic-triggers-25-increase-in-prevalence-of-anxiety-and-depression-worldwide}.
\newblock (Accessed on 11/01/2022).

\end{thebibliography}

\end{document}